\documentclass{article}

\usepackage{microtype}
\usepackage{graphicx}
\usepackage{subfigure}
\usepackage{booktabs} 
\usepackage{hyperref}

\usepackage{colortbl}
\usepackage{wrapfig}

\usepackage[preprint, nonatbib]{neurips_2025}
\usepackage[numbers]{natbib}

\usepackage[utf8]{inputenc} 
\usepackage[T1]{fontenc}  
\usepackage{hyperref}    
\usepackage{url}      
\usepackage{booktabs} 
\usepackage{amsfonts} 
\usepackage{nicefrac}  
\usepackage{microtype}
\usepackage{amsmath}
\usepackage{amssymb}
\usepackage{mathtools}
\usepackage{amsthm}

\usepackage[dvipsnames]{xcolor}

\usepackage{booktabs}
\usepackage{tabularx}

\usepackage{caption} 

\usepackage{algorithm}
\usepackage{algorithmic} 
\usepackage{multirow} 
\usepackage{titlesec}
\usepackage[capitalize,noabbrev]{cleveref}

\theoremstyle{plain}

\theoremstyle{definition}

\theoremstyle{remark}

\usepackage[textsize=tiny]{todonotes}

\DeclareMathOperator*{\mobadd}{\oplus_c}

\title{Hyperbolic Residual Quantization: Discrete Representations for Data with Latent Hierarchies}
\author{
Piotr Piękos$^{1, 2}$\thanks{The majority of this work was performed during an internship with Amazon Science.} , Subhradeep Kayal$^{1}$, Alexandros Karatzoglou$^{1}$\\[1ex]
  $^{1}$Amazon\\[1ex]
  $^{2}$KAUST, AI Initiative, Thuwal, Saudi Arabia\\[1ex]
  \texttt{piotr.piekos@kaust.edu.sa}
}

\begin{document}

\maketitle

\begin{abstract}
Hierarchical data arise in countless domains, from biological taxonomies and organizational charts to legal codes and knowledge graphs. Residual Quantization (RQ) is widely used to generate discrete, multitoken representations for such data by iteratively quantizing residuals in a multilevel codebook. However, its reliance on Euclidean geometry can introduce fundamental mismatches that hinder modeling of hierarchical branching, necessary for faithful representation of hierarchical data. In this work, we propose Hyperbolic Residual Quantization (HRQ), which embeds data natively in a hyperbolic manifold and performs residual quantization using hyperbolic operations and distance metrics. By adapting the embedding network, residual computation, and distance metric to hyperbolic geometry, HRQ imparts an inductive bias that aligns naturally with hierarchical branching.
We claim that HRQ in comparison to RQ can generate more useful for downstream tasks discrete hierarchical representations for data with latent hierarchies. We evaluate HRQ on two tasks: supervised hierarchy modeling using WordNet hypernym trees, where the model is supervised to learn the latent hierarchy - and hierarchy discovery, where, while latent hierarchy exists in the data, the model is not directly trained or evaluated on a task related to the hierarchy. Across both scenarios, HRQ hierarchical tokens yield better performance on downstream tasks compared to Euclidean RQ with gains of up to $20\%$ for the hierarchy modeling task. Our results demonstrate that integrating hyperbolic geometry into discrete representation learning substantially enhances the ability to capture latent hierarchies.

\end{abstract}

\section{Introduction}
\label{sec:introduction}

Hierarchical structures appear throughout human knowledge and information organization, serving as essential frameworks for understanding complex relationships between entities. These structures can be found in biological classifications of living organisms~\cite{mayr1968role}, business organizational structures~\cite{chandler1969strategy}, and computer file systems~\cite{mckusick1984fast}. Studies show that when children learn, they organize their knowledge in hierarchies~\cite{inhelder2013early}. This pattern extends to numerous other domains as well: from taxonomic categorization in libraries and archives to the nested organization of legal codes and regulations. Government systems typically follow hierarchical arrangements, with federal, state, and local levels each containing their own internal hierarchies. Similarly, academic disciplines are organized into fields, subfields, and specialized areas. The presence of hierarchical structures across such diverse domains reflects their importance in how humans conceptualize and organize information.

In the current state of machine learning modeling, most often continuous vectors are used to represent entities~\cite{bengio2003neural, mikolov2013efficient, DBLP:conf/nips/BordesUGWY13}. However, it can sometimes be beneficial to use discrete representations rather than continuous vector embeddings. Discrete tokens function effectively as labels because in the discrete domain, generation is equivalent to prediction. This equivalence allows models to avoid complicated generation methods like GANs~\cite{goodfellow2020generative} or diffusion models~\cite{ho2020denoising}, and instead rely on more straightforward prediction tasks. Discrete representations also tend to be more interpretable as each token can correspond to a specific concept or attribute in the hierarchy~\cite{rajput2023recommender}.

\begin{wrapfigure}{r}{0.5\linewidth}

    \centering
    \includegraphics[width=\linewidth]{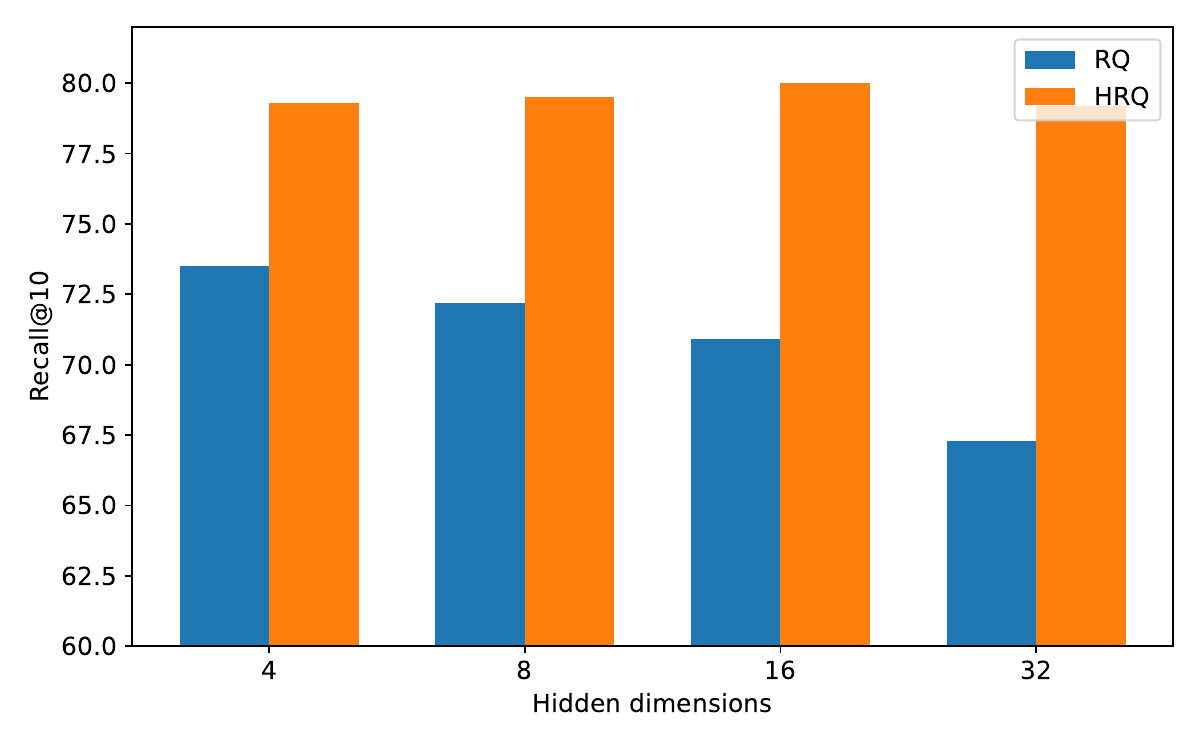}
    \caption{Recall@10 of the hypernym generation based on tokens generated by HRQ vs tokens generated by RQ. HRQ consistently outperforms RQ. Furthermore, HRQ sustains consistent scores across different dimensionalities of the embedding.}
    \label{fig:main_results}
\end{wrapfigure}

Residual Quantization Variational Autoencoders (RQ-VAE)~\cite{lee2022autoregressive, zeghidour2021soundstream} leverage these benefits by creating semantic hierarchical discrete representations through a multilevel quantization process~\cite{DBLP:conf/nips/OordVK17}. At each step of residual quantization, the model encodes increasingly fine-grained details, with earlier levels capturing broader structural elements and later levels representing more specific attributes. The result is a list of tokens that together create an identifier of the entity. We will refer to this hierarchical discrete representation as \emph{Multitoken(MT)}.  By learning discrete tokens at multiple levels of abstraction, RQ-VAE provides a framework for modeling hierarchical relations directly from dense embedding.

However, RQ-VAE operates within Euclidean space, which imposes fundamental limitations on its ability to capture hierarchical relationships. Euclidean geometry struggles to efficiently represent tree-like structures~\cite{gromov1987hyperbolic}, as the volume of space grows polynomially with distance from the origin, while the number of nodes in a hierarchy typically grows exponentially with depth. This geometric mismatch means that Euclidean-based models like RQ-VAE inevitably lose important hierarchical information during encoding. 

In contrast, hyperbolic space~\cite{gromov1987hyperbolic} - a Riemannian manifold with constant negative curvature - has been shown to model hierarchies remarkably well~\cite{nickel2017poincare, nickel2018learning}. The hyperbolic space can approximately isometrically embed any tree already in two dimensions~\cite{gromov1987hyperbolic}, whereas the same cannot be said for the Euclidean space of any dimension. The volume in the hyperbolic space grows exponentially with distance from the origin, aligning well with the growth of number of nodes in hierarchy.

The ability to encode trees by hyperbolic geometry has inspired numerous advances in machine learning. Hyperbolic neural networks have also been extensively leveraged in continuous embedding models to exploit latent hierarchies in a variety of domains. Poincaré embeddings~\cite{nickel2017poincare}, learn continuous hierarchies by mapping symbolic data into an n‑dimensional Poincaré ball. The authors showed that these embeddings outperform the Euclidean ones on tree‑structured data in terms of both representation capacity and generalization ability. Hyperbolic embeddings found immediate use in data domains rich in latent hierarchies, such as knowledge‑graph representation\cite{balazevic2019nips, chami2020acl, liang2024ecai} and recommender systems~\cite{sun2021hgcf, chamberlain2019scalable, mirvakhabova2020recsys}, and other~\cite{DBLP:journals/corr/abs-2104-11430, GaneaBH18, Liang2024}.

Despite these advances in continuous embedding models, the application of hyperbolic geometry to discrete representation learning has remained mostly underexplored. HyperVQ~\cite{goswami2024hypervq} proposes to perform vector quantization in a hyperbolic space by phrasing it as a hyperbolic multinomial logistic regression.  In this paper, we use hyperbolic distance to find the nearest codebook vector and focus on the hyperbolic version of Residual Quantization. We introduce Hyperbolic Residual Quantization (HRQ), which performs residual quantization (RQ) in a hyperbolic space with an adapted process of residual quantization to accommodate the hyperbolic structure. We claim that for \textbf{data with latent hierarchies} residual quantization benefits from hierarchical inductive bias induced by hyperbolic space. We implement this approach through several key adaptations: first, we employ hyperbolic neural networks for the embedding process, ensuring that data representations reside natively in hyperbolic space. Second, we utilize hyperbolic operations to calculate the residuals between quantization levels, preserving the geometric properties of the space throughout the quantization process. Finally, we incorporate hyperbolic distance metrics in the clustering algorithm, allowing the model to properly capture the hierarchical relationships between data points. These modifications enable HRQ-VAE to make use of the natural advantages of hyperbolic geometry to represent hierarchical structures while maintaining the benefits of discrete token-based representations.

We evaluate the quality of the multitokens created by HRQ in two scenarios. First, we test its ability to model hierarchies with supervision on the hierarchy. (H)RQ creates multitokens of nouns~\cite{miller1995wordnet} based on their hypernymy relation. Then, we test which representation is more useful in generating the hypernym for a given noun. We show that multitokens learned with Hyperbolic Residual Quantization significantly outperform tokens learned with Residual Quantization. Furthermore, we test the model's ability to create meaningful hierarchies without direct supervision on the hierarchy. Specifically, we evaluate it in a scenario where hierarchy exists, but the model is not supervised on modeling the hierarchy and is used for a task not directly related to the hierarchy. We show that the multitokens generated by HRQ outperform the multitokens generated by RQ in this scenario as well.

The paper is organized as follows. In Section~\ref{sec:background_poincare} we introduce necessary concepts from the theory of hyperbolic spaces for our method and describe the RQ-VAE algorithm. In Section~\ref{sec:method} we describe HRQ-VAE. In Section~\ref{sec:experiments} we demonstrate our experimental results. In section~\ref{sec:related_work}. Finally, in Section~\ref{sec:future_work} we summarize our findings and propose future directions.

\section{Background}

\subsection{Hyperbolic Space}

\begin{wrapfigure}{r}{0.5\linewidth}
    \centering
    \includegraphics[width=1\linewidth]{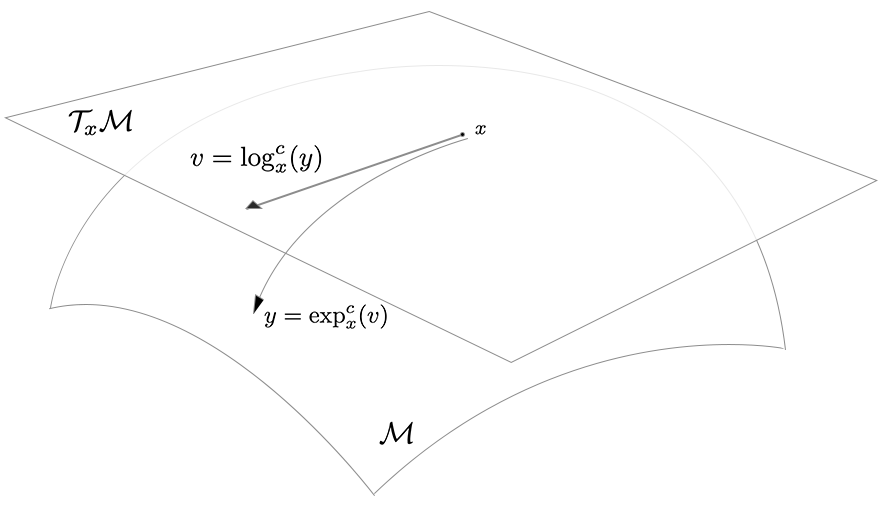}
    \caption{Visualization of the tangent space and related operations. Exponential map $exp_x^c$ maps from the tangent space attached at $x$ to the manifold and logarithmic map $log_x^c$ maps from the manifold to the tangent space attached at point $x$.}
    \label{fig:tangent}
\end{wrapfigure}

Hyperbolic geometry operates on manifolds with constant negative Gaussian curvature. A fundamental characteristic of hyperbolic geometry is its exponential spatial expansion relative to the distance from any reference point, creating abundant capacity to represent branching structures. This property enables hyperbolic spaces to accommodate the embedding of complex hierarchical relationships with minimal distortion. Research has demonstrated that arbitrary tree structures can be embedded within a hyperbolic space while approximately preserving their metric properties~\cite{gromov1987hyperbolic, hamann2018tree}. Because of these results, hyperbolic space can be conceptualized as "a continuous version of a tree," making it exceptionally valuable for computational representations of hierarchical data structures, complex networks with inherent branching patterns, and systems characterized by nested relationships.

The hyperbolic space is modeled as an open set of $\mathbb{R}^n$ with a Riemannian metric. Minkowski model uses one sheeted hyperboloid. There are, however, other models of hyperbolic space models, such as the Beltrami-Klein model or the Poincaré ball model. Both the Beltrami-Klein model and the Poincaré ball are certain projections of the Minkowski hyperboloid model. In this work, we will use the Poincaré ball model, which is the most widely used representation of the hyperbolic space in the context of neural networks. The definition of the Poincaré ball we use follows Ganea et al.~\citep{ganea2018hyperbolic}.

\paragraph{The Poincaré Ball Model.}\label{sec:background_poincare} The $n$-dimensional Poincaré Ball $\mathbb{P}_c^n$ with curvature $c$ is a set $\{x \in \mathbb{R}^n : c||x||^2 < 1\}$ with Riemannian metric $g_x^{\mathbb{P}} = \lambda_x^2 g^E$, where $g^E$ is the Euclidean metric tensor and $\lambda_x := \frac{2}{1-c||x||^2}$. The gyrovector spaces \cite{ungar2008analytic} allow one to define the operations corresponding to the standard operations in the euclidean vector spaces. In the Poincaré ball model $\mathbb{P}_c^n$ the \textbf{M\"{o}bius addition} is a hyperbolic analogue of a standard addition operation, defined as
\[ x \oplus_c y := \frac{(1+2c\langle x, y\rangle + c ||y||^2)x+(1-c||x||^2)y}{1+2c\langle x,y\rangle +c^2||x||^2||y||^2} 
 \quad\Big|\quad   x \ominus_c y :=  x \oplus _c (-y)\] 

\paragraph{Hyperbolic Distance.}

Distance in the Poincaré ball model of the hyperbolic space is defined as
\begin{equation}
    d^{\mathbb{P}_c}(u, v) = \text{arcosh}(1 + 2 \frac{c||u-v||^2}{(1-c||u||^2)(1-c||v||^2)})
    \label{eq:d_hyp}
\end{equation}

 As points get farther from the center, the distance between them grows exponentially, creating increasingly more space near the boundaries. Conversely, there is limited space near the center, naturally constraining which points can occupy these central positions - a property that aligns with hierarchical structures where few elements serve as high-level abstractions. The metric treats distance differently when moving toward/away from the center versus moving side-to-side, which helps capture both how deep items are in the hierarchy and how they branch apart. Items that belong to the same branch end up close to each other but at different depths, while the exponential growth of distances ensures effective separation between different branches. This makes it easy to preserve both the local structure (items close to each other in the hierarchy) and the overall organization (how different branches relate to each other).

\paragraph{The Tangent Space.} The tangent space $\mathcal{T}_x\mathcal{M}$ of the manifold $\mathcal{M}$ at point $x$ is an euclidean space attached to the manifold at point $x$ that intuitively contains all possible velocities the vector attached to $x$ can have. 

In order to translate between manifold and tangent space, two special maps are used. The exponential map projects vectors from the tangent space $\mathcal{T}_x\mathcal{M}$ to the manifold $\mathcal{M}$. In contrast, the logarithmic map is used to project from the manifold $\mathcal{M}$ to the tangent space $\mathcal{T}_x\mathcal{M}$. 

For the Poincaré ball model, the exponential and logarithmic maps are equal to
\begin{align*}
\exp_x^c(v)&= x \oplus_c \left(\tanh\left(\sqrt{c}\frac{\lambda_x\|v\|}{2}\right)\frac{v}{\sqrt{c}\|v\|}\right) \\
\log_x^c(y) &= \frac{2}{\sqrt{c}\lambda_x}\tanh^{-1}(\sqrt{c}\|-x \oplus_c y\|)\frac{-x \oplus_c y}{\|-x \oplus_c y\|}
\end{align*}
Similarly to the addition, scalar multiplication has its own hyperbolic version. These operations suffice to derive linear layers. Furthermore, with exponential and logarithmic maps, it is possible to add nonlinearities by translating back and forth from the manifold to the tangent space. 

Here, we defined only necessary the concepts that will be explicitly used in the HRQ-VAE algorithm, and omitted others that are necessary to derive hyperbolic layers (like scalar multiplication). We refer interested readers to \citet{ganea2018hyperbolic} or \citet{cannon1997hyperbolic}

\begin{figure*}[btp]  
    \centering
    \includegraphics[width=\textwidth]{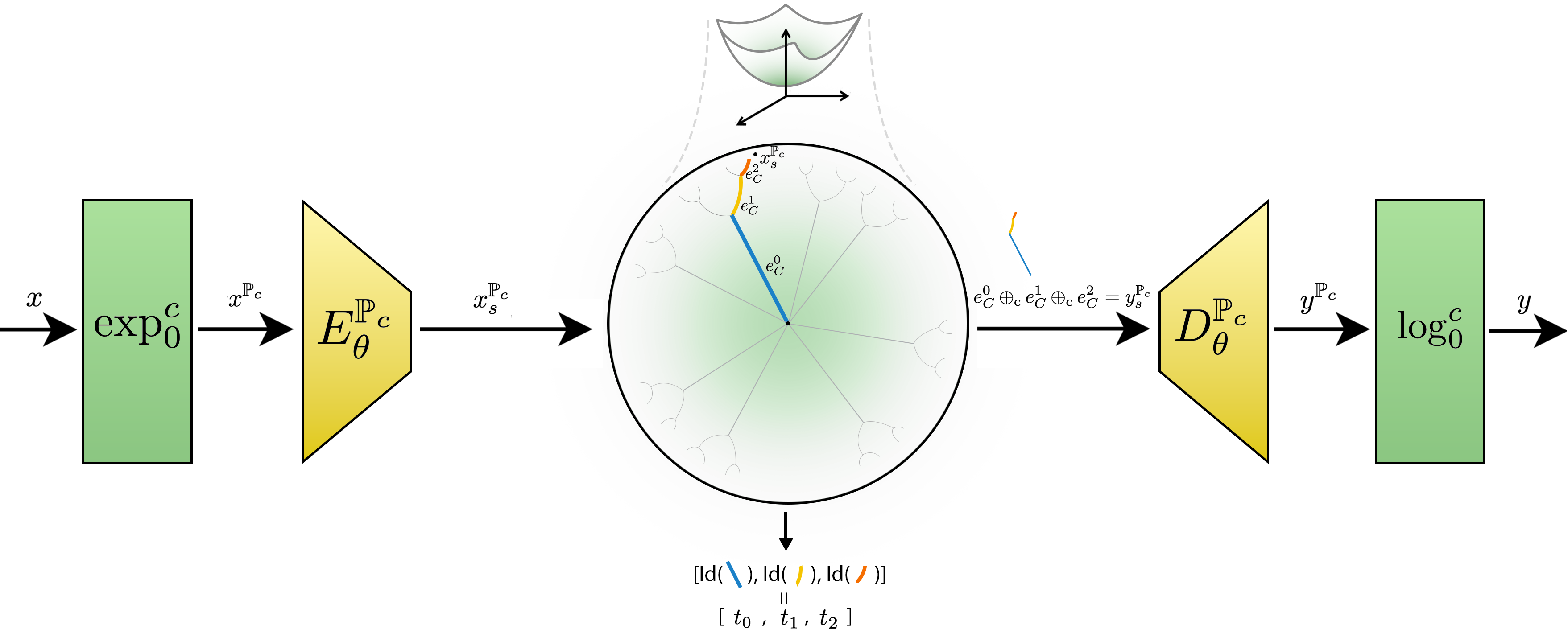}
    \caption{HRQ-VAE visualized. In the image HRQ-VAE quantizes given vector $x$ into a multitoken $[t_0, t_1, t_2]$ and its corresponding embeddings $e_C^0, e_C^1, e_C^2$. Green blocks represent mapping to and from hyperbolic space. Yellow blocks represent hyperbolic autoencoder. The detailed part in the middle is responsible for hyperbolic residual quantization. The space expands exponentially the further we go away from the center. In fact, The circle's border is at infinite distance from point 0. As a consequence, most of the points must be distant from the center and only a small number of points can be at a privileged position close to the center. This leads to natural occurence of hierarchies.  Light gray branches represent the possible HRQ-VAE and  }
    \label{fig:hrqvae_pipeline}
\end{figure*}

\subsection{Multitokens and Residual Quantization (RQ)}\label{sec:rq}
\vspace{-0.7em}

\textbf{Multitoken(MT)} is a list of discrete tokens that \emph{together} identify an entity in the dataset $D$. While the typical flat discrete representation is a single number from $0$ to $|D|-1$, the multitoken of length $k$, is a list of $k$ tokens $[t_0, ... ,t_{k-1}]$, such that jointly they identify a corresponding entity. If multitokens are structured in a semantic way, they can offer representational benefits over flat tokens. Specfically, tokens can be shared across different multitokens, leading to information sharing and a more efficient and robust representation than flat tokens, where each entity is treated independently. For example, a tiger might be identified by a multitoken [12,24] and a lion might be identified by a token [12,364]. In this case, the first token $12$ is shared between the two entities and leads to a shared part of the representation. The difficulty lies in creating good, structurally semantic multitokens.

\textbf{Residual Quantization (RQ)} approximates vectors through an iterative process that quantizes the difference (residual) between the original vector and its current approximation at each step. By sequentially quantizing these residuals using multiple codebooks, RQ produces the multitoken that represents the vector. The multitokens produced by RQ are characterized by their hierarchical structure. In general, multitokens do not have to be hierarchical. For example, each token in the multitoken might be independent. RQ due to how vector are created induces a hierarchical structure, where first tokens represent more general concepts. 

Let $C = [C_0,..., C_{k-1}]$ be a codebook, where each $C_i$ is a list of $s$ vectors from $\mathbb{R}^h$. For $j$ in ${0,...,s-1}$ we say that $C_i[j]$ is the $j$'th element of $C_i$. RQ for a vector $x_s \in \mathbb{R}^h$ produces a sequence of tokens $[t^0, t^1, ..., t^{k-1}]$ that can be associated with the corresponding codebook vectors $[e^0_C \in C_0, e^1_C \in C_1, ... , e^{k-1}_C \in C_{k-1}]$ by iteratively quantizing the residuals. RQ quantizes the vector as follows. First, the residual is set to $r^0 = x_s$. Then $r^0$ is quantized as $e^0,t^0 = q_{C_0}(r^0)$, where $e^0$ is the closest vector to the codebook $C_0$ and $t^0 \in \{0, ..., s-1\}$ is its ID: $e^0=C_0[t^0]$. After that, the residual $r^1 = r^0 - e^0$ is calculated, and we repeat the process to calculate a new token $t_1$ and its embedding $e^1$. This is repeated until we have $k$ tokens $[t^0,...t^{k-1}]$ and $k$ embeddings $[e^0_C, ..., e^{k-1}_C]$. $[t^0, t^1, ..., t^{k-1}]$ is called a multitoken of $x_s$.

The quantized embeddings are added together to create a quantized reconstruction $y_s=\sum_{i=0}^{k-1} e_C^i$. In a situation where the vectors match the codebooks, specifically when $r^{k-1} \in C_{k-1}$, $y_s = x_s$. In other cases, $|x_s-y_s|$ should be relatively small, so in order to use it in training, the derivative is modeled as an identity function $\frac{dy_s}{dx_s}=I$. We write $RQ_C(x_s) = ([t_0,..., t_{k-1}], y_s)$. 

RQ is usually combined into bigger models like RQ-VAE. Then, there is loss associated with RQ: $L_{RQ}(x_s) = \sum_{i=1}^k \big(||sg[r_C^i] - e_C^i||^2 + \alpha||r_C^i - sg[e_C^i]||^2\big)$, which is used to learn the codebook. The parameter $\alpha$ controls whether the residuals should be driven towards the current codebook or whether the codebook should be driven towards the residuals.

To solve conflicts between items in the representations generated by RQ, it adds an additional token that extends multitoken to uniquely identify the item. In practice, this is rarely necessary to uniquely identify an item as the multitokens from RQ most often suffice for the identification.

\subsection{RQ-VAE}

RQ-VAE learns the codebook vectors jointly with an autoencoder. It is an autoencoder combined with RQ process done on the latent embeddings. 

Let $E_\theta:\mathbb{R}^d \rightarrow \mathbb{R}^{d_h}$, $D_\theta:\mathbb{R}^{d_h}\rightarrow \mathbb{R}^d$ and $C$ be the codebook for RQ defined as above. $\hat{x}$ is a reconstructed $x$ that is calculated by $\hat{x} = D_\theta(y_s)$, where $([t_0, ..., t_{k-1}], y_s) = RQ_C(E_\theta(x))$. The autoencoder parameters $\theta$ are learned jointly with the quantization codebook $C$. To do that, two losses are used. Standard reconstruction loss $L_R(x) = ||x-\hat{x}||^2$ and RQ loss $L_{RQ}(x_s)$.  The reconstruction loss learns the semantic embedding and prevents the collapse of the encoded vector, whereas RQ loss learns the codebook while controlling how residuals are pushed towards current codebook as well.

\vspace{-0.5em}
\section{Method}\label{sec:method}

\vspace{-0.5em}
\subsection{Hyperbolic Residual Quantization}

Residual Quantization creates a sequential, hierarchical representation of the vector. However, when hierarchies appear in the data, the Euclidean space might structure them in a suboptimal way to learn the multitokens. On the other hand, hierarchies tend to be well-aligned in the hyperbolic space. Following that observation, we do the residual quantization in the hyperbolic space after the following adaptations. We call our new algorithm Hyperbolic Residual Quantization (HRQ).

First, we assume that the vector $x_s$ lies in the hyperbolic space. For clarity, we will call it $x_s^{\mathbb{P}_c}$. Each codebook $C_i$ is a list of $s$ vectors from $\mathbb{P}_c^{h}$. Furthermore, when selecting the closest vector $e^0_C,t_0 = q_{C}^{\mathbb{P}_c}(r^0)$ from the codebook, we use the hyperbolic distance (Eq.~\ref{eq:d_hyp}). Moreover, since we operate in the hyperbolic space, we calculate the residual with the M\"obius subtraction $r^i = r^{i-1} \ominus_c e_C^{i-1}$. The quantized reconstruction is, as a consequence, also created with the M\"obius addition \( y_s = \mobadd\limits_{i=0}^{\,k-1} e^i_{C} \). These ensure that the underlying algebraic operations preserve the hyperbolic structure of the vectors. 

The rest is analogous to RQ~(Section~\ref{sec:rq}). We write ${HRQ}_C(x_s^{\mathbb{P}_c})=([t_0,...,t_{k-1}], y_s)$

\subsection{HRQ-VAE}
HRQ-VAE, similarly to RQ-VAE, creates multitoken representations from the dense embedding. 
It learns the autoencoder jointly with the codebook. We design HRQ-VAE to operate on dense \emph{Euclidean} vectors, therefore we include exponential and logarithmic maps to transfer from the flat Euclidean space to the manifold. 

Furthermore, for encoder and decoder we use hyperbolic networks~\cite{ganea2018hyperbolic}. Let $E_\theta^{\mathbb{P}_c}: \mathbb{P}_c^h \rightarrow \mathbb{P}_c^{h_s}$,  $D_\theta^{\mathbb{P}_c}: \mathbb{P}_c^{h_s} \rightarrow \mathbb{P}_c^{h}$ and $C$ be the hyperbolic codebook. For HRQ-VAE, $\hat{x}$ is calculated as $\hat{x} = \log_0^c(D_\theta^{\mathbb{P}_c}(y_s^{\mathbb{P}_c}))$, where $([t_0,...,t_{k-1}], y_s^{\mathbb{P}_c}) = HRQ_C(E_\theta^{\mathbb{P}_c}(\exp_0^c(x))$.

The training losses are the same as for RQ-VAE. However, HRQ-VAE is trained with a Riemannian optimizer~\cite{becigneul2018riemannian}. The visualization of the HRQ-VAE is shown in the Figure~\ref{fig:hrqvae_pipeline}. The pseudocode for HRQ-VAE is in the Algorithm~\ref{alg:hrq-vae}.

\newcolumntype{Y}{>{\centering\arraybackslash}X}

\begin{table*}[hb]
\centering
{
\begin{tabularx}{\textwidth}{l l l YYYY}
\toprule
\multirow{2}{*}{$|C_i|$} & \multirow{2}{*}{$k$} & \multirow{2}{*}{Token type} & \multicolumn{4}{c}{Hidden dimensions} \\
\cmidrule(lr){4-7}
      &      &        & 4      & 8      & 16     & 32     \\
\midrule\multirow{4}{*}{64}
  & \multirow{2}{*}{3}
    & RQ  & 71.2\% & 73.7\% & 69.8\% & 67.1\% \\
  & 
    & HRQ & \textbf{79.0\%}(\textcolor{ForestGreen}{$+10.9\%$})
          & \textbf{79.6\%}(\textcolor{ForestGreen}{$+8.0\%$})
          & \textbf{79.1\%}(\textcolor{ForestGreen}{$+13.3\%$})
          & \textbf{78.3\%}(\textcolor{ForestGreen}{$+16.7\%$}) \\
\cmidrule(lr){2-7}
  & \multirow{2}{*}{4}
    & RQ  & 71.2\% & 70.9\% & 70.3\% & 64.6\% \\
  & 
    & HRQ & \textbf{78.8\%}(\textcolor{ForestGreen}{$+10.7\%$})
          & \textbf{79.2\%}(\textcolor{ForestGreen}{$+11.8\%$})
          & \textbf{79.2\%}(\textcolor{ForestGreen}{$+12.6\%$})
          & \textbf{78.9\%}(\textcolor{ForestGreen}{$+22.1\%$}) \\
\midrule
\multirow{4}{*}{128}
  & \multirow{2}{*}{3}
    & RQ  & 71.3\% & 72.5\% & 72.4\% & 66.3\% \\
  & 
    & HRQ & \textbf{79.5\%}(\textcolor{ForestGreen}{$+11.4\%$})
          & \textbf{79.5\%}(\textcolor{ForestGreen}{$+9.7\%$})
          & \textbf{79.5\%}(\textcolor{ForestGreen}{$+9.8\%$})
          & \textbf{79.1\%}(\textcolor{ForestGreen}{$+19.3\%$}) \\
\cmidrule(lr){2-7}
  & \multirow{2}{*}{4}
    & RQ  & 72.2\% & 72.7\% & 70.9\% & 64.6\% \\
  & 
    & HRQ & \textbf{79.1\%}(\textcolor{ForestGreen}{$+9.6\%$})
          & \textbf{79.4\%}(\textcolor{ForestGreen}{$+9.2\%$})
          & \textbf{79.6\%}(\textcolor{ForestGreen}{$+12.3\%$})
          & \textbf{78.7\%}(\textcolor{ForestGreen}{$+21.8\%$}) \\
\midrule
\multirow{4}{*}{256}
  & \multirow{2}{*}{3}
    & RQ  & 72.4\% & 73.2\% & 71.2\% & 66.2\% \\
  & 
    & HRQ & \textbf{78.9\%}(\textcolor{ForestGreen}{$+9.0\%$})
          & \textbf{80.0\%}(\textcolor{ForestGreen}{$+9.3\%$})
          & \textbf{80.3\%}(\textcolor{ForestGreen}{$+12.9\%$})
          & \textbf{79.9\%}(\textcolor{ForestGreen}{$+20.7\%$}) \\
\cmidrule(lr){2-7}
  & \multirow{2}{*}{4}
    & RQ  & 73.5\% & 72.2\% & 70.9\% & 67.3\% \\
  & 
    & HRQ & \textbf{79.3\%}(\textcolor{ForestGreen}{$+7.9\%$})
          & \textbf{79.5\%}(\textcolor{ForestGreen}{$+10.1\%$})
          & \textbf{80.0\%}(\textcolor{ForestGreen}{$+12.9\%$})
          & \textbf{79.2\%}(\textcolor{ForestGreen}{$+17.7\%$}) \\
\bottomrule
\end{tabularx}
}
\caption{Top 10 Recall of hypernymy prediction models trained on multitokens generated by RQ and HRQ. Despite operating on the same model and differing only in the structure of the multitokens, model that operated on HRQ multitokens produced significantly higher recall than model operating on RQ multitokens. The value (\textcolor{ForestGreen}{$+x.x\%$}) represents a percentage gain of HRQ w.r.t. RQ: (HRQ-RQ)/RQ. These results demonstrate that HRQ multitokens capture significantly more semantic information than their RQ counterparts. }
\label{tab:hierarchy_modeling_results}
\end{table*}

\section{Experiments}\label{sec:experiments}

In this section, we empirically evaluate the quality of multitokens produced by RQ and HRQ. We evaluate the quality of tokens in a two-step pipeline. First, we learn the tokens for all entities. Then, we fix the tokens and investigate how well they perform in a downstream task. 

We focus on data with latent hierarchies and our claim is that hyperbolic residual quantization produces better multitokens for data that contain latent hierarchies. Therefore, all our experiments are characterized by the clear existence of hierarchies in datasets. We evaluate HRQ in two distinct scenarios.
First, we look at Hierarchy Modeling(Section~\ref{sec:hierarchy_modeling}), in which the multitokens are explicitly trained on the hierarchy and the downstream task is related to the hierarchy multitokens are modeling. The second scenario, which we call Hierarchy Discovery(Section~\ref{sec:hierarchy_discovery}), operates a dataset which contains latent hierarchies, but the model that learns multitokens is not supervised on these hierarchies. Furthermore, the downstream task is not directly related to the latent hierarchy as well. 

Additionally, in Appendix~\ref{sec:structure} we inspect the structure of the space hypothesize on what causes benefits of HRQ multitokens. The implementation details for all methods are in Appendix~\ref{sec:implementation_details}.

\subsection{Hierarchy Modeling}\label{sec:hierarchy_modeling}

In this section, we investigate how effectively HRQ tokens capture hierarchical relationships compared to RQ. To do that, (H)RQ creates tokens by learning directly to predict hierarchical relation. Our experimental setup is similar to the main experiments from \citet{nickel2017poincare}. However, instead of operating on continuous embeddings, we compare discrete embeddings generated by Residual Quantization and Hyperbolic Residual Quantization.

Specficially, we use the transitive closure of the WordNet~\cite{miller1995wordnet} noun taxonomy. The WordNet taoxnomy contains 82,115 nouns and 743,241 hypernymy relations. A hypernymy is a semantic “is‑a” link where a general word(the hypernym) covers a group of more specific words (its hyponyms). We first learn the multitokens by simultaneously training an embedding and learning (H)RQ.

After we learn and create multitokens for all nouns, we fix multitokens, and we train a sequence-to-sequence transfomer model that translates noun to its hypernym, both represented with their corresponding multitokens. We evaluate the model by measuring recall@10 in the test dataset, which was not visible neither for the multitoken creation nor for the training of the sequence-to-sequence model. The test dataset is a randomly selected $15\%$ of all hypernymy relations. 

We learn the multitoken of the noun by embedding it in a continuous space, then contrastively pushing away nouns that are not in the hypernymy relation and pulling closer nouns that are. At the same time, the embedding is being quantized into multitokens by RQ or HRQ. Both embedding and codebook vectors are trained joinlty at the same time. 

Formally, let $N$ be the set of nouns, and $H = \{(u, v):u\in N,v\in N: \text{$u$ is a hypernym of $v$}\}$ be the set defining the hypernymy relation. Let $E_\theta$ be the $h$-dimensional embedding network, that embeds either in Euclidean or hyperbolic space depending on the model. Let $H'(u) = \{v: (u,v) \notin H\} \cup \{u\}$. We also have a (H)RQ algorithm with a codebook of length $k$, each codebook having $s$ vectors. Then the total loss for $(u,v)\in H$ is given by:
\[ L(u,v) = \log \frac{e^{-d(E_\theta(u),E_\theta(v))}}{\sum_{v'\in H'(u)} e^{-d(E_\theta(u), E_\theta(v'))}} + L_{RQ}(E_\theta(u)) + L_{RQ}(E_\theta(v))\]
In practice, we limit $H'(u)$ to 50 sample nouns from $N$ that are not in the hypernymy relation with $u$ and $u$. 

$L(u,v)$ is minimized for $\theta, C$ with $d$ being either the euclidean distance for RQ or hyperbolic distance for HRQ. As a result, it produces multitokens for all nouns $T(u) = [t_0, ..., t_{k-1}]$. In the next step, a transformer sequence-to-sequence model is trained to predict a hypernym for a given noun, both represented as their learned multitokens. The idea is that multitokens that better capture the structure of the space will serve as a more useful representation for the hypernymy generation.

To evaluate the representation quality of multitokens generated by RQ and HRQ, we investigate different combinations of parameters. We investigate the results for token lengths $k \in \{3,4\}$. We vary the size of codebooks $s \in \{64,128,256\}$ and the dimensions of dense embeddings $h \in \{4,8,16,32\}$. We focus on small dimensionalities of the dense embedding because usually before the residual quantization occurs, the embedding is mapped to a low-dimensional space.

The results for $k=4$ and $|C_i|=256$ are shown in the Figure~\ref{fig:main_results}. The complete results are shown in Table~\ref{tab:hierarchy_modeling_results}. Although the final sequence-to-sequence models differ only in the representations of the nouns and otherwise have the same architecture, the tokens generated by HRQ sometimes lead to an improvement of up to $20\%$ over the tokens generated by RQ. It clearly demonstrates the quality difference in favor of HRQ. The significant improvement is consistent across all dimensions tested. This demonstrates that HRQ is able to create significantly more semantic multitokens than the RQ, when it is trained to predict hierarchical relations.

\subsection{Hierarchy Discovery}\label{sec:hierarchy_discovery}

In real-world applications, the data often contains inherent hierarchical structures that are not explicitly labeled or available during model training. Although approaches directly supervised on the hierarchy can effectively learn to mimic known hierarchies, discovering latent hierarchical relationships without direct supervision presents a more challenging task. We call it the "Hierarchy Discovery", where the (H)RQ model creates hierarchical structures based solely on patterns present in the embeddings.

In this setting, we evaluate whether the hierarchical inductive bias of HRQ-VAE leads to multitokens that capture more semantic information compared to the standard RQ-VAE, when neither model has access to hierarchical supervision during training. Both approaches must rely entirely on patterns within the embeddings themselves, but in HRQ-VAE there is an additional inductive bias towards the formation of hierarchical structures. Our evaluation focuses on downstream task performance as the primary measure of representation quality, reflecting the practical perspective that better representations should yield improved results on real-world problems.

We use the Amazon Reviews 2014~\cite{mcauley2015image} dataset, which contains a product catalog with detailed descriptions. We will generate multitokens of the products and then use them in a recommender system. We first generate dense embeddings from product descriptions, which serve as input for both our RQ-VAE and HRQ-VAE models. We generate the embedding with an MPNet text encoder~\cite{song2020mpnet}.

RQ-VAE and HRQ-VAE are trained to produce multitokens without explicit hierarchical supervision. The quality of these discrete representations is subsequently measured by using them to train a sequence-to-sequence recommender system, where performance differences directly reflect the semantic richness captured by each quantization approach. The recommender system is a transformer encoder-decoder that predicts the next bought item based on all the previous history. Following the protocol of~\citet{rajput2023recommender} we limit the user histories to those that have at least five items and truncate the histories to 20 items. 

In order to test the model beyond the Amazon Reviews 2014 dataset, we also include the evaluation on the MovieLens 10M~\cite{harper2015movielens} dataset. Note that both product and movies can be structured in latent taxonomical hierarchies, making them suitable for our case. As the MovieLens dataset does not contain the movie description, we first generate the descriptions with the LLM Claude~\cite{claude}. The prompt used to generate the description is included in the Appendix~\ref{app:datasets}.

\begin{table}[ht]
  \centering
  \scriptsize
  \resizebox{\textwidth}{!}{%
    \begin{tabular}{@{}llccc@{}}
      \toprule
      \textbf{Dataset} & \textbf{Metric} & \textbf{Random}                & \textbf{RQ‑VAE}               & \textbf{HRQ‑VAE}              \\
      \midrule
      \multirow{4}{*}{AR Beauty}
        & NDCG@5    & 1.66\%±0.07 & 2.29\%±0.03 & \textbf{2.41\%}±0.04 (\textcolor{ForestGreen}{$+5.2\%$}) \\
        & Recall@5  & 2.06\%±0.09 & 3.68\%±0.03 & \textbf{3.74\%}±0.04 (\textcolor{ForestGreen}{$+1.6\%$})\\
        & NDCG@10   & 2.35\%±0.09 & 2.83\%±0.05 & \textbf{2.89\%}±0.05(\textcolor{ForestGreen}{$+2.1\%$}) \\
        & Recall@10 & 3.87\%±0.17 & 4.83\%±0.06 & \textbf{5.01\%}±0.06(\textcolor{ForestGreen}{$+3.7\%$}) \\
      \midrule
      \multirow{4}{*}{AR TaG}
        & NDCG@5    & 1.51\%±0.07 & 1.91\%±0.02 & \textbf{1.94\%}±0.02(\textcolor{ForestGreen}{$+1.6\%$}) \\
        & Recall@5  & 1.97\%±0.09 & 2.82\%±0.03 & \textbf{2.93\%}±0.03(\textcolor{ForestGreen}{$+3.9\%$}) \\
        & NDCG@10   & 1.94\%±0.09 & 2.45\%±0.03 & \textbf{2.47\%}±0.03(\textcolor{ForestGreen}{$+0.8\%$}) \\
        & Recall@10 & 2.76\%±0.16 & 4.22\%±0.08 & \textbf{4.53\%}±0.09(\textcolor{ForestGreen}{$+7.3\%$}) \\
      \midrule
      \multirow{4}{*}{AR SaO}
        & NDCG@5    & 0.95\%±0.07 & \textbf{1.03\%}±0.02 & \textbf{1.03\%}±0.02 ({$+0.0\%$})\\
        & Recall@5  & 1.34\%±0.08 &1.58\%±0.02 & \textbf{1.62\%}±0.02(\textcolor{ForestGreen}{$+2.5\%$}) \\
        & NDCG@10   & 1.29\%±0.09 & \textbf{1.50\%}±0.03 & 1.48\%±0.02(\textcolor{BrickRed}{$-1.4\%$})  \\
        & Recall@10 & 2.41\%±0.14 & 2.78\%±0.04 & \textbf{2.85\%}±0.04(\textcolor{ForestGreen}{$+4.0\%$})  \\
      \midrule
      \multirow{4}{*}{MovieLens}
        & NDCG@5    & 11.42\%±0.32 & 11.45\%±0.20 & \textbf{11.76\%}±0.24(\textcolor{ForestGreen}{$+2.7\%$})  \\
        & Recall@5  & 17.43\%±0.54 & 17.62\%±0.21 & \textbf{17.90\%}± 0.25(\textcolor{ForestGreen}{$+1.6\%$})  \\
        & NDCG@10   & 13.21\%±0.58 & 13.89\%±0.28 & \textbf{14.27\%}± 0.40(\textcolor{ForestGreen}{$+2.7\%$})  \\
        & Recall@10 & 23.52\%±0.73 & 25.11\%±0.37 & \textbf{25.49\%}± 0.33(\textcolor{ForestGreen}{$+1.5\%$})  \\
      \bottomrule
    \end{tabular}%
  }
  \caption{Results of recommender systems for different multitokens (Random, RQ‑VAE and HRQ‑VAE) across four datasets. The table reports average metric over 8 runs. The observed standard deviation is written on the right of the results. For the HRQ-VAE, percentage improvement over RQ-VAE is in the parantheses. }
  \label{tab:hierarchy_discovery_results}
\end{table}

Apart from the RQ-VAE and HRQ-VAE we also include a baseline that consists of randomly sampled tokens with additional token that distinguishes conflicts, similarly to (H)RQ. 
The main results are shown in Table \ref{tab:hierarchy_discovery_results}.  The multitokens generated by HRQ-VAE consistently outperform RQ-VAE and the random baseline. The reported results are on the test set with each model type selected with the highest performance on the validation set.

\section{Related Work}\label{sec:related_work}

\paragraph{Quantized representations.}

Quantized discrete representations are an alternative to dense embeddings, which recently gained popularity. The aim of a quantized representation is to create coarse information representations that focus on qualitative properties \cite{gray1984vector}. \citet{van2017neural} proposes VQ-VAE that learns the vector codebook simultaneously together with the embeddings. This was further enhanced by RQ-VAE \cite{lee2022autoregressive, zeghidour2021soundstream} that calculates the sequence of discrete tokens by iteratively quantizing the residuals. Discrete representations are beneficial to use as labels, as they avoid issues of high-dimensional continuous generation. 

VQ-GANs \cite{esser2021taming, yu2021vector} utilize vector quantization for adversarial \cite{goodfellow2020generative, schmidhuber1991possibility} image generation. \citet{zeghidour2021soundstream, yang2023hifi} uses VQ-VAE for audio generation. RQ-VAE has been introduced both in audio \cite{zeghidour2021soundstream} and image processing \cite{lee2022autoregressive}.

\paragraph{Hyperbolic Neural Networks.}

Neural networks operating in hyperbolic space have been demonstrated to perform well in tasks and modalities with hierarchical structures. \citet{sala2018representation, nickel2017poincare, ganea2018hyperboliccones} demonstrate benefits of hyperbolic embeddings for data with latent hierarchies. \citet{ganea2018hyperbolic} derives multi-layer fully connected hyperbolic neural network.  

The benefits of utilizing hyperbolic neural networks can be observed in multiple areas containing hierarchies. \citet{ma2021hyperexpan, yang2024hypformer} model the taxonomy of objects in hyperbolic space. \cite{atigh2022hyperbolic, khrulkov2020hyperbolic} applies hyperbolic nets to computer vision. Hyperbolic neural networks have shown their benefits in reinforcement learning \cite{cetin2022hyperbolic} due to the hierarchical nature of the unrolling episodes. \citet{chamberlain2019scalable, chen2022modeling, sun2021hgcf} applies hyperbolic networks to recommender systems with two-fold motivation: 1) The bipartite graph nature of the interactions between users and items, which has been shown to correspond to a complex network \cite{krioukov2010hyperbolic}, and 2) Taxonomical nature of the items. Hyper-VQ~\cite{goswami2024hypervq} proposes vector quantization in hyperbolic space.
It presents the quantization problem as a hyperbolic multinomial regression and is orthogonal to our contributions for HRQ-VAE. Both can be combined together and we consider that a promising future work.

\paragraph{Recommender Systems.}

Traditional recommender systems represent items with ID-based tokens, though with the advent of LLM usage in recommender systems, content-based tokenazation methods have been proposed. RQ-VAE has been used in recommender systems~\cite{rajput2023recommender} to tokenize item representation and train a transformer-based recommender algorithm~\cite{kang2018selfattentive}.

Another sequential generative recommendation model~\cite{petrov2023} (based on GPT-2) also applies a quantization scheme based on collaborative filtering and matrix factorization computed embedding. These embeddings are quantized to generate tokenized representations of the items. Unlike the previous approach, hierarchies are not extracted in this way. In the work of \citet{Hua2023}, a hierarchical clustering approach is presented based on graph clustering techniques of graph co-occurrence. The items are represented by the concatenation of the cluster path from the root and its own leaf token.

\section{Conclusions and future work}\label{sec:future_work}

The results shown in this work indicate that HRQ-VAE creates hierarchical representations more robust than RQ-VAE, when latent hierarchies appear in the dataset. We show that even if the model is not directly supervised on the latent hierarchy, the multitoken generated by HRQ-VAE might still be more robust than multitoken generated by RQ-VAE. Due to ubiquity of latent hierarchies in practical dataset, we see potential for number of applications of HRQ-VAE.

Furthermore, improving the performance of discrete hierarchical tokens leads to more interpretable models, as the hierarchical tokens can be related to the data taxonomies. This direction of research might lead to models whose discrete representations remain robust under domain shifts and noisy inputs, leading to societal benefits such as enhanced transparency in AI-driven decision-making, and greater public trust through auditability of the deployed systems.

In this work, we limited the scope of investigation to datasets that exhibit clear latent hierarchies. However, RQ-VAE has shown impressive results in several domains that do not follow this assumption, such as image and audio processing. HRQ-VAE, after appropriate adaptation, can potentially be applied to these domains as well. Each modality presents its own unique challenges related to the scale of experiments, hyperbolic adaptations, and the analysis of performance-contributing factors. Due to these complexities, we considered these additional modalities outside the scope of the current paper. However, exploring the application of HRQ-VAE to these diverse domains remains an exciting direction for future work.

\section*{Acknowledgements}

We would like to thank the Amazon DiscoTec Science Team, in particular Fernando Morales, Eva Mohedano Robles, David Buchaca Prats and Miriam Bellver Bueno, for the insightful discussions during the early stages of this work. We also thank Firas Laakom for his valuable comments on draft versions of the manuscript.

\bibliographystyle{plainnat}
\bibliography{example_paper}

\appendix

\section{Datasets details}
\label{app:datasets}

\subsection{Hierarchy Modeling}

WordNet is a large, manually curated lexical database of English that groups words into synonym sets (synsets) and interlinks these synsets via semantic relations such as hypernymy and hyponymy, enabling rich hierarchical modeling of concepts~\cite{miller1995wordnet}. Each synset contains a gloss (brief definition) and example usages, and synsets are organized into noun, verb, adjective, and adverb hierarchies. For our hierarchy modeling, we focus exclusively on the noun subnetwork, where the “is-a” (hypernym) relation defines a directed acyclic graph representing a noun hierarchy.

The noun subnetwork consists of $82,115$ nouns and $743,241$ hypernymy relations. We split it into the train set and test set by randomly choosing $85\%$ of the hypernymy relations to be selected for the the train set. The Embedding, RQ and the sequence-to-sequence models are all trained on the train set. We use the remaining $15\%$ as the test set on which we report the performance.

\subsection{Hierarchy Discovery}
We used four datasets to evaluate the HRQ-VAE performance in the Hierarchy Discovery section. Three data sets are the categories 'Beauty', 'Sports and Outdoors' and 'Toys and Games' from the Amazon Reviews 2014 suite \cite{mcauley2015image}. We also evaluate HRQ-VAE on the MovieLens10M dataset \cite{harper2015movielens}. 

The (H)RQ-VAE uses dense embeddings of the items to learn the corresponding hierarchical tokens. In order to create dense embeddings of the items, we use a pretrained, fixed language model embedding \cite{song2020mpnet}, which embeds the description of the item. The descriptions of the items are included in the Amazon Reviews 2014 datasets. For MovieLens, we first create the description from the movie title with the help of a Claude 3.5 Sonnet \cite{claude} language model.

\begin{table}[h]
    \centering
    \begin{tabular}{lcc}
        \toprule
        \textbf{Dataset} & \textbf{Users} & \textbf{Items} \\
        \midrule
        AR Beauty & 22,363 & 12,101 \\
        AR Toys and Games & 35,598 & 18,357 \\
        AR Sports and Outdoors & 19,412 & 11,924 \\
        MovieLens10M & 71,567 & 10,681 \\
        \bottomrule
    \end{tabular}
    \vspace{+0.5em}
    \caption{Quantitative statistics of datasets used in Hierarchy Discovery experiments.}
    \label{tab:placeholder}
\end{table}
In all experiments, we focus on predicting the next item the user interacted with (whether watched a movie or bought a product) and disregard the scores. This is a standard practice in the area of recommender systems \cite{rajput2023recommender, kang2018selfattentive, zhou2020s3}.

In order to use MovieLens, we first create the descriptions with Claude 3.5 Sonnet \cite{claude}. We use the following prompt to generate the movie description:
\begin{quote}
    \texttt{You are an expert in movie descriptions. Your task is to generate movie description that: \\
- contains a maximum of 100 words \\
- captures the general theme of the movie and interesting specifics of the story\\
- can be used adequately in a search engine to search for a movie
Your task is to generate a movie description for the following movie title. Return the movie description and do not return anything else.}
\end{quote}
From the description, we generate a dense embedding in the same way as for AR datasets. In all datasets, we cut the histories shorter than 5 elements and limit the length of user histories to 20. 

\paragraph{Test/train split.}
Following the standard evaluation \cite{rajput2023recommender} method, we divide user histories into the test, validation, and training part with a leave-one-out strategy. If the user history is a sequence of items $[i_1,...,i_T]$, with $T$ elements. The training set consists of history limited to $T-2$ tokens. The validation set is a prediction of $i_{T-1}$ based on $[i_1, ..., i_{T-2}]$ and the test set is a prediction of $i_T$ based on $[i_1,..., i_{T-1}]$. The last and second-to-last items are taken from all users for the validation and test split, regardless of the length trajectory. Note that $i_T$ in the notation above represents an item, not a token. Hence, for a multitoken scenario of tokens trained with (H)RQ-VAE, a single item $i_T$ will be represented by a multitoken of length $k$ and all $k$ atomic tokens will be selected for the test/validation set.

\section{HRQ-VAE}

\begin{algorithm}[H]
   \caption{HRQ-VAE}
   \label{alg:hrq-vae}
\begin{algorithmic}
   \STATE {\bfseries Input:} $x \in \mathbb{R}^d$

   \STATE $x^{\mathbb{P}_c} \gets \exp_0^c(x)$
   \STATE $x_s^{\mathbb{P}_c} \gets E_\theta^{\mathbb{P}_c}(x^{\mathbb{P}_c})$ 
   \STATE $y_s^{\mathbb{P}_c} \gets 0$, $r_C^0 = x^{\mathbb{P}_c}_s$
   
   \FOR{$i \in \{0,.., k-1\}$}
   \STATE $e_C^i, t_i = q_C(r_C^{i})$
   \STATE \text{add $t_i$ to the return sequence}
   \STATE $r_C^{i+1} \gets r_C^{i} \ominus_c e_c^{i-1}$
   \STATE $y_s^{\mathbb{P}_c} \gets y_s^{\mathbb{P}_c} \oplus_c e_C^i$
   \ENDFOR
   \STATE $y^{\mathbb{P}_c} \gets D_\theta^{\mathbb{P}_c}(y_s^{\mathbb{P}_c} )$
   \STATE $y \gets \log_0^c(y^{\mathbb{P}_c} )$
   \STATE $l_{\text{rec}} = || x - y ||^2$
   \STATE $l_{\text{cmt}} = \sum_{i=0}^{k-1} (||sg[r_C^i] - e_C^i||^2 + \alpha||r_C^i - sg[e_C^i]||^2)$
   \STATE $l \gets l_{\text{rec}} + l_{\text{cmt}}$
   \STATE $\nabla\theta \gets \frac{dl}{d\theta}$ ; $\nabla C \gets \frac{dl}{dC}$
   \STATE {\bfseries return} $t_0, t_1, ..., t_{k-1}, \nabla\theta, \nabla C$
\end{algorithmic}
\end{algorithm}
    \captionof{algorithm}{HRQ‑VAE}

\section{Implementation details}\label{sec:implementation_details}
\subsection{Hierarchy Modeling}

\paragraph{(H)RQ.}

To create multitokens of nouns we learn at the same time the embedding of the nouns and the codebook that quantizes the tokens. 

We investigate the results for token lengths $k \in \{3,4\}$. We vary the size of the codebooks $s \in \{64,128,256\}$ and the dimensions of dense embeddings $h \in \{4,8,16,32\}$. 
Other parameters follow \citet{nickel2017poincare}. We use Stochastic Gradient Descent~\cite{rumelhart1986learning} or Riemannian Stochastic Gradient Descent~\cite{bonnabel2013stochastic} for the optimization of encoders and RQ/HRQ codebook respectively. We use the learning rate $1.0$. We train both models for $1500$ epochs, out of which first $20$ epochs are warm-up epochs with learning rate equal to $0.01$.

\paragraph{Downstream Model.}

The sequence-to-sequence model is trained to generate hypernyms of a noun, both represented as multitokens. Hence, both the input and the output of the model are a list of $k$ tokens from $0$ to $s$. The transformer model has $4$ layers for both the encoder and the decoder. The hidden dimension is equal to $256$ with the feedforward dimension equal to $1024$ and $8$ attention heads. The embeddings of the encoder and decoder are tied. It is trained for $100$ epochs with Adam~\cite{kingma2014adam} optimizer with a learning rate equal to $0.001$.

\subsection{Hierarchy Discovery}

\paragraph{(H)RQ-VAE.} 

The initial dense embedding of the text is calculated with 768 dimensional MPNET \cite{song2020mpnet}. From the dense embedding, we train the (H)RQ-VAE and assign the new hierarchical token produced by the model to each item. 
The encoder in (H)RQ-VAE has 3 intermediate layers of size 512, 256, 128 with (H)ReLU activation and the output layer of size 32. The decoder has symmetric architecture to the encoder. The codebook has length 256 for each token and is not shared across tokens. We use batch size of 128, and train the (H)RQ-VAE for 5000 epochs with learning rates $[10^{-3}, 10^{-4}, 10^{-5}]$. We choose the learning rate that performed the best on the validation split of the downstream task and report the corresponding test result.

\paragraph{Downstream Model.}
We train the recommender system to evaluate the quality of the discrete representations produced by (H)RQ-VAE. User history is a sequence of items the user interacted with: either a movie they watched, or an item they bought. At each step, we predict the next item the user will interact with; specifically, we generate $k \in \{5, 10\}$ ranked guesses. To evaluate the quality of the set of guesses, we use two most popular recommender system metrics: Recall@K and NDCG@K.

\begin{wrapfigure}{r}{0.5\linewidth}

    \centering
    \includegraphics[width=\linewidth]{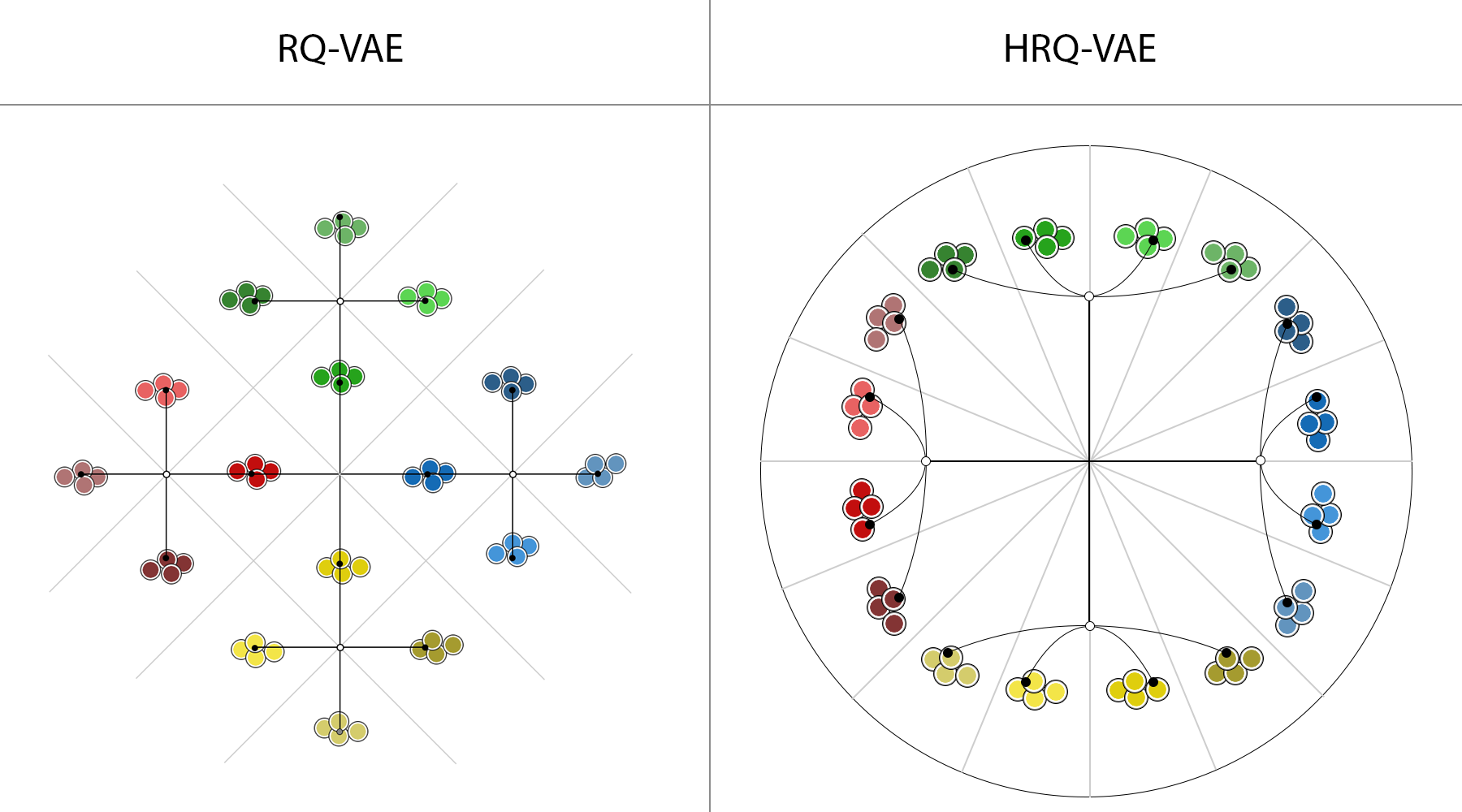}
    \caption{The embedding space structure induced by RQ-VAE and HRQ-VAE, respectively for a hierarchical tokens of length 2. The data is represented by coloured dots. Hue of the dot represents first hierarchical token. The shade represents second token. For the RQ-VAE the result is a typical effect of hierarchical clustering. HRQ-VAE due to exponential growth of the space has inductive bias to putting leaf nodes away on a similar distance away from the center.}
    \label{fig:structure}
\end{wrapfigure}

In our case of multitokens, each item is represented by a multitoken. Each user history has concanted multitokens of all items given a user bought(or a movie watched), and each specific recommendation is considered good if the entire multitoken corresponds to the true item the user interacted with. We split all the datasets into train,validation, and test set in the same way. We limit the histories to 20 interactions and filter the histories with less than 5 interactions. Furthermore, we select the last interaction as a test set, the second to last as a validation set, and everything else as a training set. 

We train a sequence-to-sequence transformer model \cite{vaswani2017attention} with T5 \cite{raffel2020exploring} architecture. For each datapoint, an output sequence is the hierarchical representation of the next item, whereas the input is all their previous history. The model has a token embedding size of 384, 6 attention heads with 64 dimension each. and 1024 dimension of the feedforward net.

Our setup for hierarchy discovery follows the parameters of \citet{rajput2023recommender}. However, the results differ significantly on the AR dataset. The fact that they differ consistently across all tokens and also across random baselines suggests that the cause of the inconsistency must lie in the final recommender system. However, after a detailed inspection and testing of different libraries, we were unable to reproduce the original results. However, please note that, contrary to \citet{rajput2023recommender} our claim is not about creating the best recommender system, but about comparing HRQ to RQ, and if the shift in the performance is caused by the downstream model - it is not important for our claim, as we use recommender system only as a downstream task to evaluate the quality of HRQ multitokens in comparison to RQ multitokens.

All experiments were ran on a device equiped in a single 16GB Nvidia-V100 card.

\vspace{-0.5em}
\section{Structure of The Space}\label{sec:structure}

\begin{wraptable}{r}{0.42\textwidth}
    \centering
    \renewcommand{\arraystretch}{1.3}
    \begin{tabular}{|l|c|c|}
        \hline
        \rowcolor{gray!15}
        \cline{2-3}
        \rowcolor{gray!15}
        & \textbf{RQ-VAE} & \textbf{HRQ-VAE} \\
        \hline
        \textbf{Variable} & $\|x_s\|_2$ & $\|\log_0^c(x_s^{\mathbb{P}_c})\|_2$ \\
        \hline
        \rowcolor{gray!5}
        \textbf{EV} & {0.7213} & {0.3251} \\
        \hline
        \textbf{Std. dev} & {0.2696} & {0.0664} \\
        \hline
        \rowcolor{gray!5}
        \textbf{CV} & 0.3738 & {0.2042} \\
        \hline
    \end{tabular}
    \caption{ Analysis of the norms for RQ-VAE and HRQ-VAE. We compare the euclidean norm of the low dimensional vector $x_s$ to the euclidean norm of hyperbolic $x_s^{\mathbb{P}_c}$ after mapping to the tangent space with logarithmic map. For the comparison we use the Coefficient of Variation defined as $CV(X)=\frac{\sigma(X)}{\mu(X)}$. It is used to compare the variability of a random variables with different orders of magnitude. RQ-VAE has almost twice the CV of HRQ-VAE which supports our claim about the structure of their corresponding spanning trees.}
    \label{tab:coefficient_variation}
\end{wraptable}

Suppose we have a set of points $S$ and we want to find a point that minimizes average distance to all points from $S$. In Euclidean space this point will be the center of mass of points, a simple average of all points from $S$. However, in hyperbolic space, the point that minimizes the average hyperbolic distance (Eq. \ref{eq:d_hyp}) will be a continuous analogue to the nearest common ancestor node of all the nodes.

This leads to a vastly different structures when these spaces are clustered hierarchically and, as a consequence, to a vastly different structures for spanning trees of the residual quantization. In the Euclidean space the points corresponding to the leafs will be splattered around the space with the qunatization tree cutting into the centers of respective subclusters. Meanwhile, in the hyperbolic case, the leafs will be mostly spread around with the cluster "centers" being closer to $0$ than the cluster points. This behavior has been observed in the hyperbolic clustering \cite{chami2020trees}. We visualize the structural differences in Fig. \ref{fig:structure}.

This structure is beneficial for learning hierarchical relations for several reasons. Because the space is split radially most of the time the regions can have their own infinite part of the space, whereas in Euclidean division some regions are crammed close to center. As a consequence, the edges between regions are sharper than in the hyperbolic space, which might lead to poorer generalization. Finally, the structure imposed by the hierarchical euclidean quantization leads to strong utilization of the vector norms to select the cluster. On the other hand, hyperbolic quantization that leads to leafs being set the most outward in the spanning tree leaves the norm for the optimizer to choose, which can be an important benefit for gradient-based learning. 

We argue that these structural difference of the hierarchical space of HRQ-VAE in comparison to space of RQ-VAE leads to the superior performance of HRQ-VAE in downstream tasks.

To quantitatively support this argument we inspect the norms of low-dimensional encoded representations $x_s$ and $x_s^{\mathbb{P}_c}$. Specifically, we argue that the norms will vary less in the hyperbolic space. To make a fair comparison we compare Euclidean norms, so the hyperbolic $x_s^{\mathbb{P}_c}$ vector is first transformed to the tangent space with logarithmic map. Moreover, as the models differ in the average norm we look at the Coefficient of Variation as a measure of interest. The coefficient of variation is defined for positive variables as $CV(X)=\frac{\sigma(X)}{\mu(X)}$. The results are shown in Table \ref{tab:coefficient_variation} and confirm that the norms vary significantly more for the vectors to be quantized in the euclidean space.

\section{Limitations}\label{sec:limitations}

Although HRQ and HRQ-VAE demonstrate better performance in the discussed tasks, they come with some limitations. The biggest limitation is the strong assumptions about the type of data. Currently, we limit the claim to the situation where the dataset has latent hierarchies, and at the same time, we are interested in discrete representations. This is a very specific situation. Extending the evaluation to domains of general application in which RQ-VAE succeeded, such as image or audio, would greatly increase the influence. However, the current version does not investigate performance in this direction. Furthermore, increase in popularity of continuous methods for generation, such as diffusion models, hinders the applicability of the generative methods based on discrete tokens.

\end{document}